\documentclass[11pt]{article}
\usepackage{coling2020}
\usepackage{times}
\usepackage{url}
\usepackage{latexsym}

\usepackage{microtype}
\colingfinalcopy 

\usepackage{amsmath}
\usepackage{multirow}
\usepackage{graphicx}
\usepackage{amssymb}
\usepackage{courier}
\usepackage{tablefootnote}

\newcommand{\model}{\textsc{PoD}}

\title{\model: Positional Dependency-Based Word Embedding \center{for Aspect Term Extraction}}

\author{Yichun Yin$^{1}$, Chenguang Wang$^{2}$, Ming Zhang$^{3}$\thanks{\hspace{1pt} Corresponding author.}\\ 
	$^{1}$ Noah's Ark Lab, Huawei\\
	$^{2}$ Computer Science Division, UC Berkeley\\
	$^{3}$ Department of Computer Science, Peking University\\
	{\tt yinyichun@huawei.com,chenguangwang@berkeley.edu} \\
	{\tt mzhang\_cs@pku.edu.cn}}

\date{}

\date{}

\begin{document}
\maketitle

\begin{abstract}
	Dependency context-based word embedding jointly learns the representations of word and dependency context, and has been proved effective in aspect term extraction. In this paper, we design the positional dependency-based word embedding (\model) which considers both dependency context and positional context for aspect term extraction. Specifically, the positional context is modeled via relative position encoding. Besides, we enhance the dependency context by integrating more lexical information (e.g., POS tags) along dependency paths. Experiments on SemEval 2014/2015/2016 datasets show that our approach outperforms other embedding methods in aspect term extraction. 

\end{abstract}

\section{Introduction}
\blfootnote{
    %
    %
    %
    %
    \hspace{-0.65cm}  
    This work is licensed under a Creative Commons 
    Attribution 4.0 International Licence.
    Licence details:
    \url{http://creativecommons.org/licenses/by/4.0/}.
    
    %
}
Aspect term extraction aims to extract expressions that represent properties of products or services from online reviews~\cite{hu2004miningb,hu2004mining,popescu2007extracting,Liu10}.
Understanding the context between words in reviews, such as through conditional random fields~\cite{pontiki2014semeval,pontiki2015semeval,pontiki-EtAl:2016:SemEval}, is the key to superior results in aspect term extraction.
Word embeddings are effective to capture the contextual information across a wide range of NLP tasks~\cite{tai-socher-manning:2015:ACL-IJCNLP,lei-barzilay-jaakkola:2015:EMNLP,BojanowskiGJM17,DevlinCLT19}. However, they only produce moderate results in aspect term extraction. Recent studies (e.g., ~\newcite{yin2016unsupervised}) indicate that this is due to the distributed nature of the word embedding~\cite{mikolov2013distributed}, which ignores the rich context between the words, such as syntactic information.

In this paper, we propose positional dependency-based word embedding (\model) to enhance the context modeling capability for aspect term extraction. \model\ explicitly captures two types of contexts, {\em dependency context} and {\em positional context}. Inspired by the simple-yet-effective position encoding in Transformer~\cite{vaswani2017attention}, \model\ models the positional context via relative position encoding~\cite{ShawUV18} between words within a fixed window. Besides, the dependency context is defined as the dependency path as well as the attached lexical information (e.g., POS tags and words) along the path. Moreover, \model\ is able to incorporate more lexical information into the semantic compositional model via the dependency context, making representations of dependency paths more informative than the ones that only consider grammatical information~\cite{yin2016unsupervised}. We then linearly combine the dependency and positional context to produce the positional dependencies among words. We also define a margin-based ranking loss to efficiently optimize \model.

Our contributions are two-fold, (\expandafter{\romannumeral1}) we propose positional dependency-based word embedding \model, which incorporates both positional context and dependency context, (\expandafter{\romannumeral2}) we compare \model\ with existing aspect term extraction methods and demonstrate that \model\ yields improved results on aspect term extraction datasets.

\begin{figure*}
	\centering
	\includegraphics[width=0.6\textwidth]{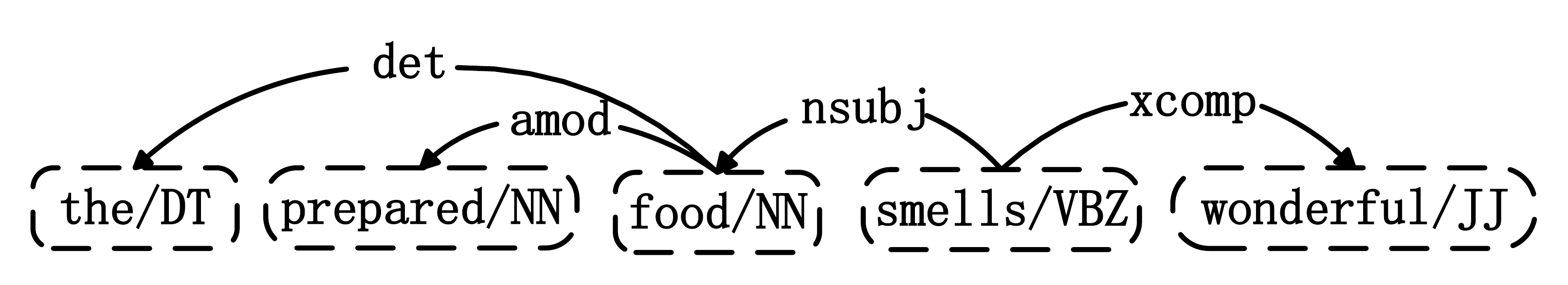}
	\vspace{-0.1in}
	\caption{An example sentence, parsed by Stanford CoreNLP.}
	\label{fig:parsing_tree}
\end{figure*}

\section{Positional Dependency-Based Word Embedding}

\subsection{Model Description}
\label{subsec:model}
\model\ aims to maximize likelihoods of triples ($\mathrm{w_t}$, $\mathrm{c}$, $\mathrm{w_c}$), where $\mathrm{w_t}$ and $\mathrm{w_c}$ represent target word and context word respectively, $\mathrm{c}$ refers to positional dependency-based context (an example is in Table~\ref{label:triples}), which consists of two types of contexts: the dependency context (dependency paths between target and context word) and positional context (relative position encoding between target and context word). Figure~\ref{fig:parsing_tree} illustrates the sentence example according to the triples in Table~\ref{label:triples}. 

We introduce two score functions for triples ($\mathrm{w_t}$, $\mathrm{c}$, $\mathrm{w_c}$) which are as follows.
\begin{equation}
\label{eq:obj}
 S_{add} = (\textbf{w}_c+ \textbf{c})\cdot \textbf{w}_t^\intercal; \\
 S_{puct} = (\textbf{w}_c \circ \textbf{c}) \cdot \textbf{w}_t^\intercal,
\end{equation}
where $S_{add}$ uses the element-wise addition for the context word and its context $\mathrm{c}$, while $S_{puct}$ uses the element-wise product. We use two embedding matrices $\textbf{M}_{t}  \in R^{|V|\times d}$ and $\textbf{M}_{c}  \in R^{|V|\times d}$ to represent target words and context words respectively, where $|V|$ is the size of vocabulary and $d$ is the dimension of embeddings. The $\textbf{w}_c \in R^{1\times d}$ and $\textbf{w}_t \in R^{1\times d}$ are obtained through lookup operations. Note that we describe how to derive $\textbf{c}$ in Section 2.2.


\begin{table}
	\centering 
	\scalebox{0.8}{
		\begin{tabular}{|c|c|c|c|}
			\hline
			\textbf{Target} &\textbf{Context} & DC & PC \\
			\hline
			\hline
			\multirow{4}{*}{\textbf{\large food}} & the & $*\stackrel{det}{\longrightarrow}*$ & -2  \\
			& prepared & $*\stackrel{amod}{\longrightarrow}*$ & -1 \\
			& smells & $*\stackrel{nsubj}{\longleftarrow}*$ & 1 \\
			& wonderful & $*\stackrel{nsubj}{\longleftarrow} \mathrm{smells}/\mathrm{VBZ}  \stackrel{xcomp}{\longrightarrow}* $ & 2\\
			\hline
		\end{tabular}
	}
	\vspace{-0.1in}
	\caption{Target word, context words and their corresponding contexts: DC refers to dependency context and PC refers to positional context.}
	\label{label:triples}
\end{table}

\vspace{-0.05in}
\subsection{Positional Dependency}
\vspace{-0.05in}
We construct the positional dependency-based context $\textbf{c}$ by linearly combining the dependency context vector $\textbf{c}_{dep}$ derived from semantic composition of lexical dependency paths and the positional context vector $\textbf{c}_{pos}$ computed based on relative position encoding~\cite{ShawUV18}. The representation of positional dependency-based context is defined in Eq.~(\ref{eqn:r}).
\begin{equation}
\textbf{c} = \alpha \cdot \textbf{c}_{pos} + (1 - \alpha) \cdot \textbf{c}_{dep}, 
\label{eqn:r}
\end{equation}
where $\alpha$ is used to trade-off the effects between dependency and positional contexts in the model.

The basic idea of using relative position encoding is based on the assumption that context words with different relative positions have different impacts on learning the representations of target words. The use of relative position encoding has been proved to be useful in supervised relation classification~\cite{zeng2014relation} and machine translation~\cite{vaswani2017attention,ShawUV18}. Similar to using embeddings to represent words, we also introduce $\textbf{M}_{l}  \in R^{(s - 1) \times d}$ to represent the relative position encoding and derive $\textbf{c}_{pos}$ from it, where $s$ is the window size.

We also consider the lexical information along dependency paths when learning the representations of the dependency context. For example, for the pair ($\mathrm{food}$, $\mathrm{wonderful}$) in Figure~\ref{fig:parsing_tree}, the corresponding dependency path is $*\stackrel{nsubj}{\longleftarrow} \mathrm{smells}/\mathrm{VBZ}  \stackrel{xcomp}{\longrightarrow} * $. 
We denote the words, POS tags as the lexical information, and use $dep = \{g_1, g_2, ..., g_{|c|}\}$ to denote the composite lexical dependency path. The embedding matrix $\textbf{M}_{dep} \in R^{n \times d} $ is utilized to derive the distributed representations of lexical dependency path $\{\textbf{g}_1, \textbf{g}_2, ..., \textbf{g}_{|c|}\}$, where $n$ is the size of dictionary including words, POS tags and dependency paths. To obtain $\textbf{c}_{dep}$, we use RNN model which learns the dependency path representations along the sequence $dep$ in a recurrent manner.

\vspace{-0.05in}
\subsection{Model Optimization}
\vspace{-0.05in}
We use a margin-based ranking objective to learn model parameters in Eq.~(\ref{eq:obj}), which encourages scores of positive triples $( \mathrm{w_t}, \mathrm{c}, \mathrm{w_c}) \in \mathcal{T}$ to be higher than scores of sampled triples $( \mathrm{w_t'}, \mathrm{c}, \mathrm{w_c}) \in \mathcal{T'}$. The ranking loss is as follows.
\begin{equation}
\begin{split}
L = \sum_{( \mathrm{w_t}, \mathrm{c}, \mathrm{w_c}) \in \mathcal{T}} \sum_{(\mathrm{w_t'}, \mathrm{c}, \mathrm{w_c}) \in \mathcal{T'}} max \{S(\mathrm{w_t}, \mathrm{c}, \mathrm{w_c}) -
		 S(\mathrm{w_t'}, \mathrm{c}, \mathrm{w_c}) + \delta, 0\},
\end{split}
\label{objective}
\end{equation}	
where $\delta$ is the margin value, $S(*)$ is the score function defined in Eq.~(\ref{eq:obj}), in which \textbf{c} is introduced in Eq.~(\ref{eqn:r}).

Note that, the proposed Eq.~(\ref{objective}) conducts negative sampling on target words rather than dependency paths, which proposes two advantages, (\expandafter{\romannumeral1}) it can exploit arbitrary hop dependency paths. Besides, the words and POS tags along the path can be utilized; (\expandafter{\romannumeral2}) it avoids to memorize dependency path frequencies which grow exponentially with the number of hops.

The negative sampling method is employed to train the embedding model (Eq.~(\ref{eq:obj})). These randomly chosen words in $\mathcal{T'}$ are sampled based on the marginal distribution $p(w)$ and $p(w)$ is estimated from the word frequency raised to the $\frac{3}{4}$ power~\cite{mikolov2013efficient} in the corpus. We set the negative number to 15 which is a trade-off between the training time and performance. The $\delta$ is empirically set to 1 according to~\cite{collobert2008unified,bollegala-maehara-kawarabayashi:2015:ACL-IJCNLP}.
To avoid the overfitting in RNN, we employ dropout on the input vectors and set the dropout rate to 0.5. The asynchronous gradient descent is used for parallel training. Moreover, Adagrad~\cite{duchi2011adaptive} is used to adaptively change learning rate and the initial learning rate is set to 0.1.

\vspace{-0.05in}
\section{Experiment}
\vspace{-0.05in}
\subsection{Dataset}
\vspace{-0.05in}
We evaluate \model\ on aspect term extraction benchmark datasets: SemEval 2014/2015/2016. The SemEval 2014 datasets include two domains: laptop and restaurant, and we use the D1 and D2 to denote these two datasets respectively. The SemEval 2015/2016 datasets only include restaurant domain. D3 and D4 are utilized to represent them. We use the corpora introduced in~\cite{yin2016unsupervised} to learn the distributed representations of words and lexical dependency paths. 



\vspace{-0.05in}
\subsection{Baseline and Setting}
\vspace{-0.05in}
We compare \model\ with top systems in SemEval with method class \emph{Top system} as shown in Table 1. We also compare our method with notable embedding-based methods with method class \emph{Embedding method} illustrated in Table 1.

In order to choose \textit{l}, \textit{d} (Section~\ref{subsec:model}) and $\alpha$ (Eq.~(\ref{eqn:r})), 80\% sentences in training data are used as training set, and the rest 20\% are used as development set. The dimensions of word and dependency path embeddings are set as 100. Larger dimensions get similar results in the development set but cost more time. \textit{l} is set as 10 which performs best in the development set. Similarly, the $\alpha$s are set as 0.7, 0.5, 0.5 and 0.5 for datasets D1, D2, D3 and D4 respectively. 

To make fair comparisons, we choose parameters \textit{l} and \textit{d} on the development set for embedding baselines. All the dimensions of embedding methods are set as 100. The dimensions $l$ in Skip-gram, CBOW and WDEmb models are set as 15, the dimensions in Glove and DepEmb are set as 10. The windows of Skip-gram, CBOW and Glove are set as 5, which are the same as our model. As derived embeddings are not necessarily in a bounded range~\cite{turian2010word}, this might lead to moderate results. We apply a simple function of discretization following~\cite{yin2016unsupervised} to make embedding features more effective.

\begin{table}
	\centering
	\scalebox{0.70}{
	\begin{tabular}{|l|l|c|c|c|c|}
		\hline
		\textbf{Method} & \textbf{Method Class} & \textbf{D1} & \textbf{D2} & \textbf{D3} & \textbf{D4} \\
		\hline
		\hline
		IHS\_RD~\cite{chernyshevich2014ihs} & Top system in D1 & \textbf{74.55} & 79.62 & - & - \\
		DLIREC~\cite{zhiqiang2014dlirec} & Top system in D2 & 73.78 & 84.01 & - & - \\
		EliXa~\cite{sanvicente-saralegi-agerri:2015:SemEval} & Top system in D3 & - & - & 70.04 & - \\
		Nlangp~\cite{toh-su:2016:SemEval} & Top system in D4 & - & - & - & \textbf{72.34} \\
		\hline
		\hline
		DRNLM~\cite{mirowski2015dependency} & Embedding method & 66.91 & 78.59 & 64.75 & 63.89 \\
		Skip-gram~\cite{mikolov2013distributed} & Embedding method & 70.52 & 82.20 & 66.98 & 68.57 \\
		CBOW~\cite{mikolov2013efficient} & Embedding method & 69.80 & 81.98 & 67.09 & 67.43\\
		Glove~\cite{pennington-socher-manning:2014:EMNLP2014} & Embedding method & 67.23 & 80.69  & 64.12 & 64.39 \\
		DepEmb~\cite{levy2014dependency} & Embedding method & 71.02 & 82.78 & 67.55 & 69.23\\
		WDEmb~\cite{yin2016unsupervised} & Embedding method & 73.72 & 83.52 & 68.27 & 70.20\\
		\hline
		\hline
		Ours-\model\ ($S_{add}$) & Embedding method & 73.54$^\ast$ & 84.21$^\dagger$ & 69.14$^\ast$ & 70.90$^\dagger$ \\
		Ours-\model\ ($S_{puct}$) & Embedding method & {74.07}$^\ast$ & \textbf{84.82}$^\ast$ & \textbf{70.18}$^\dagger$ & 71.70$^\ast$ \\
		\hline
	\end{tabular}
	}
	\vspace{-0.1in}
	\caption{Comparison of F1 scores on the SemEval 2014/2015/2016 datasets. In t-tests, the marker $^\ast$ refers to p-value $<$ 0.05, the marker $^\dagger$ refers to p-value $<$ 0.01, and WDEmb~\cite{yin2016unsupervised} is the compared method.}
	\label{embs_results}
\end{table}

\vspace{-0.05in}
\subsection{Result and Analysis}
\vspace{-0.05in}

The results are described in Table~\ref{embs_results} and the t-test is also conducted by random initialization. From the table, we find that \model\ with both $S_{puct}$ and $S_{add}$ consistently outperform WDEmb which is one of the best embedding methods. The reasons are that (\expandafter{\romannumeral1}) our model incorporates positional context as relative position encoding to help enhance word embeddings; (\expandafter{\romannumeral2}) the dependency context leverages the lexical dependency path capturing more specific lexical information such as words and POS tags (extracted using Stanford CoreNLP~\cite{ManningSBFBM14}) than WDEmb.
\model\ also achieves comparable results with top systems which are based on hand-crafted features in all datasets, which shows that our learned embeddings are effective for aspect term extraction. 
The $S_{puct}$ performs better than $S_{add}$, which indicates that the product-based composition method is more capable in capturing the useful features in aspect term extraction.
In terms of embedding-based baselines, DepEmb and WDEmb perform better than other baselines, which indicates that encoding syntactic knowledge into word embeddings is desirable for aspect term extraction. 


We also analyze the effects of POS tags and words along dependency paths in the dependency context on final results. The results are presented in Table~\ref{lexical_results}.
From the table, we observe that both POS tags and words along dependency paths boost aspect term extraction, which indicates that lexical information can encode discriminative information for representations of dependency paths. Meanwhile, \model\ obtains better results by adding both POS tags and words.  
\begin{table}
	\centering
	\scalebox{0.85}{
	\begin{tabular}{|l|c|c|c|c|}
		\hline
		\textbf{Information} & \textbf{D1} & \textbf{D2} & \textbf{D3} & \textbf{D4} \\
		\hline
		\hline 
		Dependency path & 72.13 & 83.52 & 68.39 & 70.90 \\
		+ POS tags (only) & 72.48  & 83.87 & 69.03 & 71.02 \\
		+ Words (only) & 73.79 & 84.31 & 69.98 & 71.24 \\
		+ POS tags + Words & \textbf{74.07} & \textbf{84.82} & \textbf{70.18} & \textbf{71.70} \\
		\hline
	\end{tabular}
	}
	\vspace{-0.1in}
	\caption{Effects of information in dependency context.}
	\label{lexical_results}
\end{table}



\vspace{-0.05in}
\section{Related Work}
\vspace{-0.05in}

Association rule mining is used in \cite{hu2004mining} to mine aspect terms. Opinion words are used to extract infrequent aspect terms. The relationship between opinion words and aspect words is crucial to extract aspect terms, which are deployed in many follow-up studies. In \cite{qiu2011opinion}, the predefined dependency paths are utilized to iteratively extract aspect terms and opinion words. \model\ instead learns the representation of the dependency context.

Dependency-based word embedding~\cite{levy2014dependency,KomninosM16} encodes dependencies into word embeddings, and has been shown effective in aspect term extraction as well~\cite{yin2016unsupervised}.
However, only grammatical information is considered among the dependency paths.
We instead introduce a positional dependency-based embedding method which considers both dependency context and positional context.
End-to-end aspect term extraction~\cite{DBLP:conf/emnlp/WangPDX16,DBLP:conf/aaai/WangPDX17,ijcai2018-583,xu_acl2018} based on neural networks and attention mechanism, have been recently developed. Compare to these methods, \model\ is an embedding method, can thus be applied to more applications.
Compare to deep word representations~\cite{PetersNIGCLZ18,DevlinCLT19,wang190409408}, \model\ is more efficient, which is crucial to aspect term extraction.
Text-to-network~\cite{WangSERZH15,WangSLZH15,WangSLZH16,WangSSHSWZ16,WangSRZH16,WangSLSZ017,WangSLZH18} is in general relevant to aspect term extraction, we focus on proposing a more light weighted embedding method.

\vspace{-0.05in}
\section{Conclusion}
\vspace{-0.05in}

In this paper, we develop a specific word embedding method for aspect term extraction. Our method considers both positional and dependency context when learning the word embedding. Meanwhile, the lexical information along dependency path is encoded into representations of dependency context. Compared with other embedding methods, our method achieves better results in aspect term extraction. We plan to apply our method to more NLP tasks~\cite{WangDZZ13,WangSRWHJZ15,WangACLXX17,WangCL17}.


\section{Acknowledgement}
This paper is supported by National Key Research and Development Program of China with Grant No. 2018AAA0101900 / 2018AAA0101902 as well as the National Natural Science Foundation of China (NSFC Grant No. 61772039 and No. 91646202). Chenguang Wang is supported by Berkeley DeepDrive and Berkeley Artificial Intelligence Research.

\bibliography{coling2020}

\begin{thebibliography}{}

\bibitem[\protect\citename{Bojanowski \bgroup et al.\egroup
  }2017]{BojanowskiGJM17}
Piotr Bojanowski, Edouard Grave, Armand Joulin, and Tomas Mikolov.
\newblock 2017.
\newblock Enriching word vectors with subword information.
\newblock {\em {TACL}}, 5:135--146.

\bibitem[\protect\citename{Bollegala \bgroup et al.\egroup
  }2015]{bollegala-maehara-kawarabayashi:2015:ACL-IJCNLP}
Danushka Bollegala, Takanori Maehara, and Ken-ichi Kawarabayashi.
\newblock 2015.
\newblock Unsupervised cross-domain word representation learning.
\newblock In {\em Proceedings of the 53rd Annual Meeting of the Association for
  Computational Linguistics and the 7th International Joint Conference on
  Natural Language Processing (Volume 1: Long Papers)}, pages 730--740, July.

\bibitem[\protect\citename{Chernyshevich}2014]{chernyshevich2014ihs}
Maryna Chernyshevich.
\newblock 2014.
\newblock Ihs r\&d belarus: Cross-domain extraction of product features using
  conditional random fields.
\newblock {\em SemEval 2014}, page 309.

\bibitem[\protect\citename{Collobert and Weston}2008]{collobert2008unified}
Ronan Collobert and Jason Weston.
\newblock 2008.
\newblock A unified architecture for natural language processing: Deep neural
  networks with multitask learning.
\newblock In {\em ICML}, pages 160--167.

\bibitem[\protect\citename{Devlin \bgroup et al.\egroup }2019]{DevlinCLT19}
Jacob Devlin, Ming{-}Wei Chang, Kenton Lee, and Kristina Toutanova.
\newblock 2019.
\newblock {BERT:} pre-training of deep bidirectional transformers for language
  understanding.
\newblock In {\em NAACL}, pages 4171--4186.

\bibitem[\protect\citename{Duchi \bgroup et al.\egroup
  }2011]{duchi2011adaptive}
John Duchi, Elad Hazan, and Yoram Singer.
\newblock 2011.
\newblock Adaptive subgradient methods for online learning and stochastic
  optimization.
\newblock {\em Journal of Machine Learning Research}, 12(Jul):2121--2159.

\bibitem[\protect\citename{Hu and Liu}2004a]{hu2004miningb}
Minqing Hu and Bing Liu.
\newblock 2004a.
\newblock Mining and summarizing customer reviews.
\newblock In {\em SIGKDD}, pages 168--177.

\bibitem[\protect\citename{Hu and Liu}2004b]{hu2004mining}
Minqing Hu and Bing Liu.
\newblock 2004b.
\newblock Mining opinion features in customer reviews.
\newblock In {\em AAAI}, volume~4, pages 755--760.

\bibitem[\protect\citename{Komninos and Manandhar}2016]{KomninosM16}
Alexandros Komninos and Suresh Manandhar.
\newblock 2016.
\newblock Dependency based embeddings for sentence classification tasks.
\newblock In {\em NAACL}, pages 1490--1500.

\bibitem[\protect\citename{Lei \bgroup et al.\egroup
  }2015]{lei-barzilay-jaakkola:2015:EMNLP}
Tao Lei, Regina Barzilay, and Tommi Jaakkola.
\newblock 2015.
\newblock Molding cnns for text: non-linear, non-consecutive convolutions.
\newblock In {\em EMNLP}, pages 1565--1575, September.

\bibitem[\protect\citename{Levy and Goldberg}2014]{levy2014dependency}
Omer Levy and Yoav Goldberg.
\newblock 2014.
\newblock Dependency-based word embeddings.
\newblock In {\em ACL}, pages 302--308.

\bibitem[\protect\citename{Li \bgroup et al.\egroup }2018]{ijcai2018-583}
Xin Li, Lidong Bing, Piji Li, Wai Lam, and Zhimou Yang.
\newblock 2018.
\newblock Aspect term extraction with history attention and selective
  transformation.
\newblock In {\em Proceedings of the Twenty-Seventh International Joint
  Conference on Artificial Intelligence, {IJCAI-18}}, pages 4194--4200.
  International Joint Conferences on Artificial Intelligence Organization, 7.

\bibitem[\protect\citename{Liu}2010]{Liu10}
Bing Liu.
\newblock 2010.
\newblock Sentiment analysis and subjectivity.
\newblock In {\em Handbook of Natural Language Processing, Second Edition.},
  pages 627--666.

\bibitem[\protect\citename{Manning \bgroup et al.\egroup }2014]{ManningSBFBM14}
Christopher~D. Manning, Mihai Surdeanu, John Bauer, Jenny~Rose Finkel, Steven
  Bethard, and David McClosky.
\newblock 2014.
\newblock The stanford corenlp natural language processing toolkit.
\newblock In {\em ACL}, pages 55--60.

\bibitem[\protect\citename{Mikolov \bgroup et al.\egroup
  }2013a]{mikolov2013efficient}
Tomas Mikolov, Kai Chen, Greg Corrado, and Jeffrey Dean.
\newblock 2013a.
\newblock Efficient estimation of word representations in vector space.
\newblock {\em arXiv preprint arXiv:1301.3781}.

\bibitem[\protect\citename{Mikolov \bgroup et al.\egroup
  }2013b]{mikolov2013distributed}
Tomas Mikolov, Ilya Sutskever, Kai Chen, Greg~S Corrado, and Jeff Dean.
\newblock 2013b.
\newblock Distributed representations of words and phrases and their
  compositionality.
\newblock In {\em NIPS}, pages 3111--3119.

\bibitem[\protect\citename{Mirowski and Vlachos}2015]{mirowski2015dependency}
Piotr Mirowski and Andreas Vlachos.
\newblock 2015.
\newblock Dependency recurrent neural language models for sentence completion.
\newblock {\em arXiv preprint arXiv:1507.01193}.

\bibitem[\protect\citename{Pennington \bgroup et al.\egroup
  }2014]{pennington-socher-manning:2014:EMNLP2014}
Jeffrey Pennington, Richard Socher, and Christopher Manning.
\newblock 2014.
\newblock Glove: Global vectors for word representation.
\newblock In {\em EMNLP}, pages 1532--1543, Doha, Qatar, October.

\bibitem[\protect\citename{Peters \bgroup et al.\egroup }2018]{PetersNIGCLZ18}
Matthew~E. Peters, Mark Neumann, Mohit Iyyer, Matt Gardner, Christopher Clark,
  Kenton Lee, and Luke Zettlemoyer.
\newblock 2018.
\newblock Deep contextualized word representations.
\newblock In {\em NAACL}, pages 2227--2237.

\bibitem[\protect\citename{Pontiki \bgroup et al.\egroup
  }2014]{pontiki2014semeval}
Maria Pontiki, Haris Papageorgiou, Dimitrios Galanis, Ion Androutsopoulos, John
  Pavlopoulos, and Suresh Manandhar.
\newblock 2014.
\newblock Semeval-2014 task 4: Aspect based sentiment analysis.
\newblock In {\em SemEval 2014}, pages 27--35.

\bibitem[\protect\citename{Pontiki \bgroup et al.\egroup
  }2015]{pontiki2015semeval}
Maria Pontiki, Dimitrios Galanis, Haris Papageogiou, Suresh Manandhar, and Ion
  Androutsopoulos.
\newblock 2015.
\newblock Semeval-2015 task 12: Aspect based sentiment analysis.
\newblock In {\em Proceedings of the 9th International Workshop on Semantic
  Evaluation (SemEval 2015), Denver, Colorado}.

\bibitem[\protect\citename{Pontiki \bgroup et al.\egroup
  }2016]{pontiki-EtAl:2016:SemEval}
Maria Pontiki, Dimitris Galanis, Haris Papageorgiou, Ion Androutsopoulos,
  Suresh Manandhar, Mohammad AL-Smadi, Mahmoud Al-Ayyoub, Yanyan Zhao, Bing
  Qin, Orphee De~Clercq, Veronique Hoste, Marianna Apidianaki, Xavier Tannier,
  Natalia Loukachevitch, Evgeniy Kotelnikov, N\'{u}ria Bel, Salud~Mar\'{i}a
  Jim\'{e}nez-Zafra, and G\"{u}l\c{s}en Eryi\u{g}it.
\newblock 2016.
\newblock Semeval-2016 task 5: Aspect based sentiment analysis.
\newblock In {\em Proceedings of the 10th International Workshop on Semantic
  Evaluation (SemEval-2016)}, pages 19--30, June.

\bibitem[\protect\citename{Popescu and Etzioni}2007]{popescu2007extracting}
Ana-Maria Popescu and Orena Etzioni.
\newblock 2007.
\newblock Extracting product features and opinions from reviews.
\newblock In {\em Natural language processing and text mining}, pages 9--28.

\bibitem[\protect\citename{Qiu \bgroup et al.\egroup }2011]{qiu2011opinion}
Guang Qiu, Bing Liu, Jiajun Bu, and Chun Chen.
\newblock 2011.
\newblock Opinion word expansion and target extraction through double
  propagation.
\newblock {\em Computational linguistics}, 37(1):9--27.

\bibitem[\protect\citename{San~Vicente \bgroup et al.\egroup
  }2015]{sanvicente-saralegi-agerri:2015:SemEval}
I\~{n}aki San~Vicente, Xabier Saralegi, and Rodrigo Agerri.
\newblock 2015.
\newblock Elixa: A modular and flexible absa platform.
\newblock In {\em Proceedings of the 9th International Workshop on Semantic
  Evaluation (SemEval 2015)}, pages 748--752, June.

\bibitem[\protect\citename{Shaw \bgroup et al.\egroup }2018]{ShawUV18}
Peter Shaw, Jakob Uszkoreit, and Ashish Vaswani.
\newblock 2018.
\newblock Self-attention with relative position representations.
\newblock In {\em NAACL-HLT}, pages 464--468.

\bibitem[\protect\citename{Tai \bgroup et al.\egroup
  }2015]{tai-socher-manning:2015:ACL-IJCNLP}
Kai~Sheng Tai, Richard Socher, and Christopher~D. Manning.
\newblock 2015.
\newblock Improved semantic representations from tree-structured long
  short-term memory networks.
\newblock In {\em Proceedings of the 53rd Annual Meeting of the Association for
  Computational Linguistics and the 7th International Joint Conference on
  Natural Language Processing (Volume 1: Long Papers)}, pages 1556--1566, July.

\bibitem[\protect\citename{Toh and Su}2016]{toh-su:2016:SemEval}
Zhiqiang Toh and Jian Su.
\newblock 2016.
\newblock Nlangp at semeval-2016 task 5: Improving aspect based sentiment
  analysis using neural network features.
\newblock In {\em Proceedings of the 10th International Workshop on Semantic
  Evaluation (SemEval-2016)}, pages 287--293, June.

\bibitem[\protect\citename{Turian \bgroup et al.\egroup }2010]{turian2010word}
Joseph Turian, Lev Ratinov, and Yoshua Bengio.
\newblock 2010.
\newblock Word representations: a simple and general method for semi-supervised
  learning.
\newblock In {\em ACL}, pages 384--394.

\bibitem[\protect\citename{Vaswani \bgroup et al.\egroup
  }2017]{vaswani2017attention}
Ashish Vaswani, Noam Shazeer, Niki Parmar, Jakob Uszkoreit, Llion Jones,
  Aidan~N Gomez, {\L}ukasz Kaiser, and Illia Polosukhin.
\newblock 2017.
\newblock Attention is all you need.
\newblock In {\em NIPS}, pages 5998--6008.

\bibitem[\protect\citename{Wang \bgroup et al.\egroup }2013]{WangDZZ13}
Chenguang Wang, Nan Duan, Ming Zhou, and Ming Zhang.
\newblock 2013.
\newblock Paraphrasing adaptation for web search ranking.
\newblock In {\em ACL}, pages 41--46.

\bibitem[\protect\citename{Wang \bgroup et al.\egroup }2015a]{WangSERZH15}
Chenguang Wang, Yangqiu Song, Ahmed El{-}Kishky, Dan Roth, Ming Zhang, and
  Jiawei Han.
\newblock 2015a.
\newblock Incorporating world knowledge to document clustering via
  heterogeneous information networks.
\newblock In {\em SIGKDD}, pages 1215--1224.

\bibitem[\protect\citename{Wang \bgroup et al.\egroup }2015b]{WangSLZH15}
Chenguang Wang, Yangqiu Song, Haoran Li, Ming Zhang, and Jiawei Han.
\newblock 2015b.
\newblock Knowsim: {A} document similarity measure on structured heterogeneous
  information networks.
\newblock In {\em ICDM}, pages 1015--1020.

\bibitem[\protect\citename{Wang \bgroup et al.\egroup }2015c]{WangSRWHJZ15}
Chenguang Wang, Yangqiu Song, Dan Roth, Chi Wang, Jiawei Han, Heng Ji, and Ming
  Zhang.
\newblock 2015c.
\newblock Constrained information-theoretic tripartite graph clustering to
  identify semantically similar relations.
\newblock In {\em IJCAI}, pages 3882--3889.

\bibitem[\protect\citename{Wang \bgroup et al.\egroup }2016a]{WangSLZH16}
Chenguang Wang, Yangqiu Song, Haoran Li, Ming Zhang, and Jiawei Han.
\newblock 2016a.
\newblock Text classification with heterogeneous information network kernels.
\newblock In {\em AAAI}, pages 2130--2136.

\bibitem[\protect\citename{Wang \bgroup et al.\egroup }2016b]{WangSRZH16}
Chenguang Wang, Yangqiu Song, Dan Roth, Ming Zhang, and Jiawei Han.
\newblock 2016b.
\newblock World knowledge as indirect supervision for document clustering.
\newblock {\em {TKDD}}, 11(2):13:1--13:36.

\bibitem[\protect\citename{Wang \bgroup et al.\egroup }2016c]{WangSSHSWZ16}
Chenguang Wang, Yizhou Sun, Yanglei Song, Jiawei Han, Yangqiu Song, Lidan Wang,
  and Ming Zhang.
\newblock 2016c.
\newblock Relsim: Relation similarity search in schema-rich heterogeneous
  information networks.
\newblock In {\em SDM}, pages 621--629.

\bibitem[\protect\citename{Wang \bgroup et al.\egroup
  }2016d]{DBLP:conf/emnlp/WangPDX16}
Wenya Wang, Sinno~Jialin Pan, Daniel Dahlmeier, and Xiaokui Xiao.
\newblock 2016d.
\newblock Recursive neural conditional random fields for aspect-based sentiment
  analysis.
\newblock In {\em EMNLP}, pages 616--626.

\bibitem[\protect\citename{Wang \bgroup et al.\egroup }2017a]{WangACLXX17}
Chenguang Wang, Alan Akbik, Laura Chiticariu, Yunyao Li, Fei Xia, and Anbang
  Xu.
\newblock 2017a.
\newblock {CROWD-IN-THE-LOOP:} {A} hybrid approach for annotating semantic
  roles.
\newblock In {\em EMNLP}, pages 1913--1922.

\bibitem[\protect\citename{Wang \bgroup et al.\egroup }2017b]{WangCL17}
Chenguang Wang, Laura Chiticariu, and Yunyao Li.
\newblock 2017b.
\newblock Active learning for black-box semantic role labeling with neural
  factors.
\newblock In {\em IJCAI}, pages 2908--2914.

\bibitem[\protect\citename{Wang \bgroup et al.\egroup }2017c]{WangSLSZ017}
Chenguang Wang, Yangqiu Song, Haoran Li, Yizhou Sun, Ming Zhang, and Jiawei
  Han.
\newblock 2017c.
\newblock Distant meta-path similarities for text-based heterogeneous
  information networks.
\newblock In {\em CIKM}, pages 1629--1638.

\bibitem[\protect\citename{Wang \bgroup et al.\egroup
  }2017d]{DBLP:conf/aaai/WangPDX17}
Wenya Wang, Sinno~Jialin Pan, Daniel Dahlmeier, and Xiaokui Xiao.
\newblock 2017d.
\newblock Coupled multi-layer attentions for co-extraction of aspect and
  opinion terms.
\newblock In {\em AAAI}, pages 3316--3322.

\bibitem[\protect\citename{Wang \bgroup et al.\egroup }2018]{WangSLZH18}
Chenguang Wang, Yangqiu Song, Haoran Li, Ming Zhang, and Jiawei Han.
\newblock 2018.
\newblock Unsupervised meta-path selection for text similarity measure based on
  heterogeneous information networks.
\newblock {\em Data Min. Knowl. Discov.}, 32(6):1735--1767.

\bibitem[\protect\citename{Wang \bgroup et al.\egroup }2019]{wang190409408}
Chenguang Wang, Mu~Li, and Alexander~J. Smola.
\newblock 2019.
\newblock Language models with transformers.
\newblock {\em CoRR}, abs/1904.09408.

\bibitem[\protect\citename{Xu \bgroup et al.\egroup }2018]{xu_acl2018}
Hu~Xu, Bing Liu, Lei Shu, and Philip~S. Yu.
\newblock 2018.
\newblock Double embeddings and cnn-based sequence labeling for aspect
  extraction.
\newblock In {\em ACL}.

\bibitem[\protect\citename{Yin \bgroup et al.\egroup
  }2016]{yin2016unsupervised}
Yichun Yin, Furu Wei, Li~Dong, Kaimeng Xu, Ming Zhang, and Ming Zhou.
\newblock 2016.
\newblock Unsupervised word and dependency path embeddings for aspect term
  extraction.
\newblock {\em IJCAI}.

\bibitem[\protect\citename{Zeng \bgroup et al.\egroup }2014]{zeng2014relation}
Daojian Zeng, Kang Liu, Siwei Lai, Guangyou Zhou, Jun Zhao, et~al.
\newblock 2014.
\newblock Relation classification via convolutional deep neural network.
\newblock In {\em COLING}, pages 2335--2344.

\bibitem[\protect\citename{Zhiqiang and Wenting}2014]{zhiqiang2014dlirec}
Toh Zhiqiang and Wang Wenting.
\newblock 2014.
\newblock Dlirec: Aspect term extraction and term polarity classification
  system.

\end{thebibliography}
\bibliographystyle{coling}

\end{document}